\PassOptionsToPackage{numbers,compress}{natbib}
\PassOptionsToPackage{table,dvipsnames}{xcolor}

\documentclass{article}
\usepackage[preprint]{neurips_2026}

\usepackage[utf8]{inputenc}
\usepackage[T1]{fontenc}
\usepackage{graphicx}
\usepackage{booktabs}
\usepackage{amsmath,amsfonts,amssymb}
\usepackage{hyperref}
\usepackage{url}
\usepackage{xcolor}
\usepackage{cleveref}
\usepackage{multirow}
\usepackage{array}
\usepackage{makecell}
\usepackage{pifont}
\usepackage{enumitem}
\usepackage{xspace}
\usepackage{caption}
\usepackage{wrapfig}
\usepackage{microtype}
\usepackage{tcolorbox}

\newcounter{algorithm}

\graphicspath{{./}{figures/}{tables/}}

\newcommand{\ATMA}{A-TMA\xspace}

\newcommand{\Graphiti}{Graphiti/Zep\xspace}
\newcommand{\Ariadne}{AriadneMem\xspace}

\newcommand{\targacc}{Acc\xspace}
\newcommand{\evidsupport}{Evidence support\xspace}
\newcommand{\bankcoexist}{Bank coexist\xspace}
\newcommand{\bankrole}{Bank role\xspace}
\newcommand{\cmark}{\ding{51}}
\newcommand{\xmark}{\ding{55}}

\definecolor{atmaheader}{RGB}{204,225,229}
\definecolor{atmarow}{gray}{0.94}
\definecolor{atmahlight}{RGB}{223,253,253}
\definecolor{atmagreen}{RGB}{0,128,80}
\definecolor{atmared}{RGB}{180,35,24}
\definecolor{atmablue}{RGB}{40,90,160}
\definecolor{lightblue}{RGB}{64,135,190}
\newcommand{\yesmark}{\textcolor{atmagreen}{\cmark}}
\newcommand{\nomark}{\textcolor{atmared}{\xmark}}

\newcommand{\updelta}[1]{\ensuremath{{}_{\textcolor{atmared}{\scriptscriptstyle\uparrow#1}}}}
\newcommand{\downdelta}[1]{\ensuremath{{}_{\textcolor{atmablue}{\scriptscriptstyle\downarrow#1}}}}
\newenvironment{atmaalgo}{\par\smallskip\noindent\begin{minipage}{\columnwidth}}{\end{minipage}\par\smallskip}

\title{A-TMA: Decoupling State-Aware Memory Failures in Long-Term Agent Memory}
\author{
  Zitong Shi\textsuperscript{1} \qquad
  Yixuan Tang\textsuperscript{1}\thanks{Corresponding author.} \qquad
  Anthony Kum Hoe Tung\textsuperscript{1}\\[0.7em]
  \normalfont\textsuperscript{1}National University of Singapore\\
  \normalfont\texttt{zitongshi@u.nus.edu, yixuan@comp.nus.edu.sg, atung@comp.nus.edu.sg}
}

\begin{document}

\maketitle

\begin{abstract}
Long term memory lets LLM agents act as persistent assistants, but user facts change. The central question is not only whether a fact can be remembered, but whether the system can answer from the state the user is asking about: what is true now, what used to be true, and what changed. We study \emph{ghost memory}, a state coordination failure in which old, current, and transition facts coexist in the memory bank, remain mixed during retrieval, and mislead the answer model. The challenge is that simple deletion loses history, while timestamps or relevance scores alone do not expose which fact should be live for a given query. We therefore argue for decoupled memory evaluation at three levels: bank maintenance, retrieval, and answer time resolution. We propose \ATMA, a state aware overlay for existing memory systems that keeps superseded and transition records, builds evidence packets for the requested state view, and exposes current, historical, and transition labels to QA. To make this failure measurable, we build LTP (LoCoMo Temporal Plus), a conflict heavy benchmark for ghost memory, and evaluate on LoCoMo for long conversation generalization. On LTP, \Graphiti+\ATMA improves conflict accuracy by 0.240 absolute over \Graphiti. On LoCoMo, \Graphiti+\ATMA raises temporal F1 from 0.0295 to 0.1705. The gains are host dependent, but they indicate that explicit state roles can reduce memory failures hidden by final QA accuracy.
\end{abstract}

\section{Introduction}
LLM agents are moving from single turn chat toward persistent assistance. They remember user preferences, project decisions, constraints, and earlier conversations, then reuse this context when later requests depend on it. Agent systems already use memory for simulation, reflection, tool use, and open ended interaction~\citep{park2023generative,yao2022react,shinn2023reflexion,wang2023voyager,wang2023surveyagents}. Long term memory is therefore becoming infrastructure for agent systems~\citep{chhikara2025mem0,xu2025amem,rasmussen2025zep,zhong2023memorybank,packer2023memgpt,wang2024memoryllm,li2025memos}. Yet user memory is not a static profile. A user may change an address, leave a job, revise a plan, or ask about a past state. A memory system that treats every stored fact as timeless can be wrong even when each fact was true when it was written.

We call this failure \emph{ghost memory}. Ghost memory is not the mere presence of an old record. It is a state coordination failure. Old, current, and transition facts coexist in the bank, remain mixed during evidence construction, and reach the answer model without a clear state role. A former address should answer a question about where the user lived before. The same address should not answer where a package should be shipped now. A transition note can explain what changed. It should not be treated as the live value. The core challenge is therefore query conditioned state alignment: the system must preserve history, decide which state the query asks for, and expose that state to retrieval and QA. We study the following problem: \emph{how can state changing memory be evaluated by decoupling bank, retrieval, and QA failures while preserving state history, retrieving evidence for the requested state view, and resolving answers with explicit state roles?}

\begin{figure*}[t]
    \centering
    \includegraphics[width=0.94\linewidth]{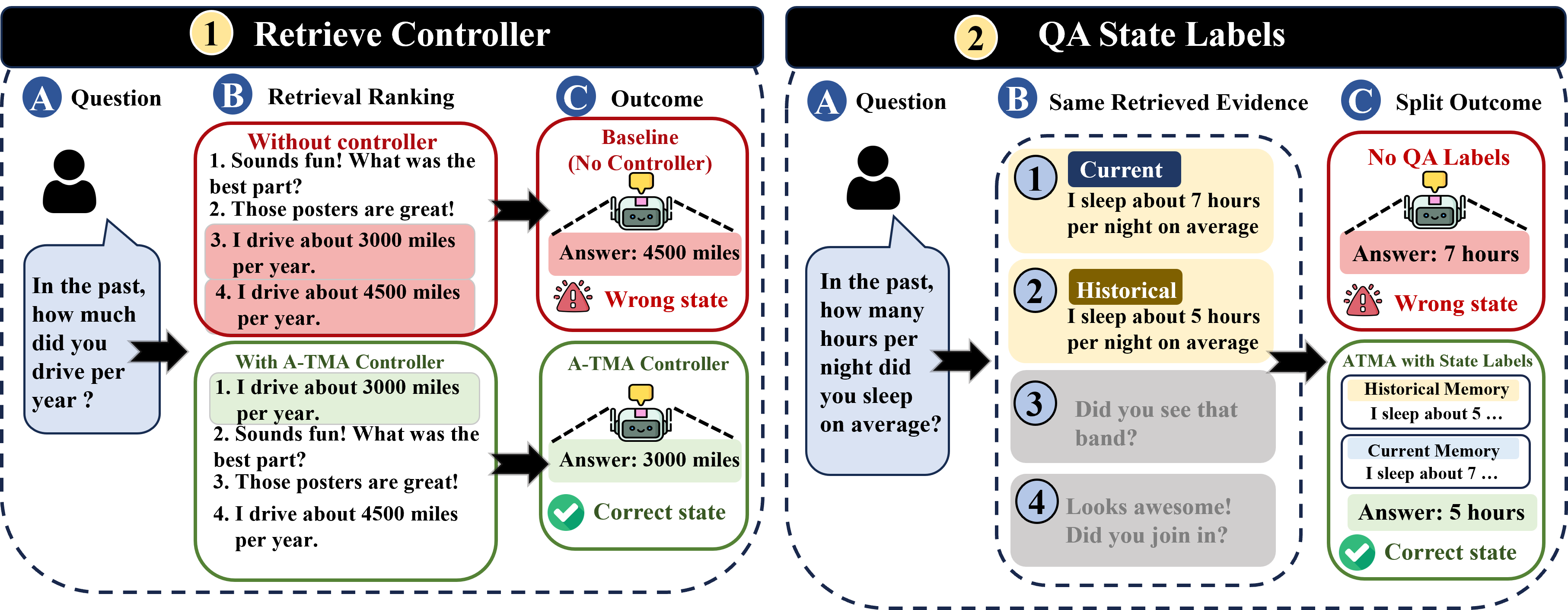}
    \caption{\textbf{Retrieval and QA state alignment.} A historical query can retrieve both old and current values, but an uncontrolled packet may foreground the wrong state. Even with the right evidence present, unlabeled context can make QA follow the current value. \ATMA uses bounded retrieval control and explicit state labels to answer from the requested state.}
    \label{fig:intro}
\end{figure*}

Most long term memory systems address only part of this requirement~\citep{chhikara2025mem0,xu2025amem,rasmussen2025zep,pan2025secom,wang2023longmem,wang2025mplus,zhu2026ariadnemem,jimenez2024hipporag,huo2026atommem}. Profile and note based systems extract user facts and retrieve them as personal context, but often treat relevance as sufficient for reuse. Temporal and episodic systems add timestamps, sessions, or recency cues. These signals help order events, but they do not by themselves expose an actionable role such as active, superseded, or transition evidence for retrieval and QA. Graph and controller based systems add entities, relations, paths, or learned write and retrieve policies, but state roles can still remain implicit across the bank, evidence packet, and final answer. Ghost memory is therefore not only a storage problem. It couples bank maintenance, retrieval, and QA.

\begin{wrapfigure}{r}{0.55\columnwidth}
    \vspace{-4pt}
    \centering
    \includegraphics[width=\linewidth]{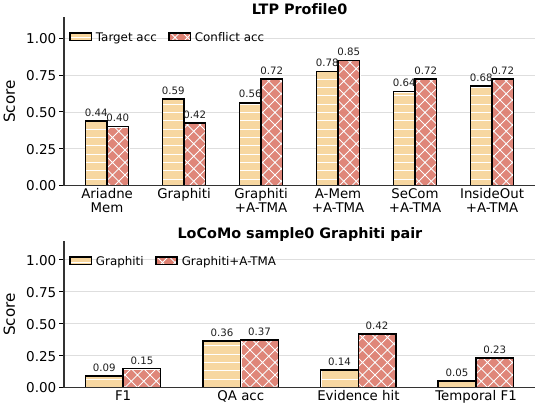}
    \vspace{-15pt}
    \captionsetup{font=small}
    \caption{\small \textbf{Subset Statistics.} Timestamped or graph labeled evidence can still fail at state correct QA. Explicit state roles reduce this failure in LTP and improve the Graphiti pair on LoCoMo sample0.}
\label{fig:timestamp-state-subset-stats}
    \vspace{-14pt}
\end{wrapfigure}

This challenge matters because metadata that looks state aware may still fail at answer time. We compare timestamped or graph labeled evidence with explicit state role evidence in Figure~\ref{fig:timestamp-state-subset-stats}. On LTP profile000, \Ariadne~\citep{zhu2026ariadnemem} provides timestamped graph facts and reasoning paths but reaches 0.438 \targacc. A-Mem+\ATMA reaches 0.775, InsideOut+\ATMA reaches 0.675, and SeCom+\ATMA reaches 0.637 on the same subset. The paired \Graphiti~\citep{rasmussen2025zep,zepai2024graphiti} result isolates the issue more directly. Temporal KG evidence reaches 0.812 evidence support but only 0.425 conflict accuracy, while \Graphiti+\ATMA raises conflict accuracy to 0.725 by exposing state roles to QA. On LoCoMo~\citep{maharana2024locomo} sample0, the effect is more metric dependent. \Graphiti+\ATMA improves over \Graphiti in F1, QA accuracy, evidence hit rate, and temporal F1, while \Ariadne remains stronger in F1. The bounded conclusion is that timestamps and graph labels help identify relevant evolving facts, but explicit state roles are still needed when retrieval and QA must decide which state the query asks for.

This also creates an evaluation problem. A final answer can be wrong for different reasons. The memory bank may fail to preserve old and current states cleanly. Retrieval may choose evidence for the wrong state view. The answer model may see the needed evidence and still collapse current, historical, and transition facts into one response. A single QA score hides these cases and makes it hard to tell whether a method failed at storage, retrieval, or answer time resolution.

We propose \textbf{A}daptive \textbf{T}emporal \textbf{M}emory \textbf{A}lignment (\ATMA), a state aware overlay for existing memory systems. \ATMA does not replace the host memory substrate. It adds state semantics around the host pipeline so that each level can be inspected. At the bank level, \ATMA keeps historical memories, marks superseded facts, and records supersession or transition links. This avoids destructive overwriting while preventing obsolete facts from staying live by default. At the retrieval level, \ATMA infers the query state view and builds a state aligned evidence packet from host retrieval results, state metadata, relation expansion, and bounded controller ranking. At the QA level, \ATMA serializes evidence with labels such as current memory, historical memory, transition memory, and transition linked memory, then asks the answer model to resolve the state requested by the query.

\looseness = -1
We evaluate \ATMA on two benchmark settings. We build LTP (LoCoMo Temporal Plus) as a conflict heavy benchmark for ghost memory, with 10 profiles and 800 probes. LoCoMo~\citep{maharana2024locomo} tests long conversation generalization over 10 samples and 1,986 QA pairs. On LTP, \Graphiti+\ATMA improves conflict accuracy by 0.240 absolute, from 0.480 to 0.720. On LoCoMo, \Graphiti+\ATMA raises temporal F1 from 0.0295 to 0.1705 and improves average F1 from 0.0809 to 0.1556. The gains are host and metric dependent, with the clearest benefit when the host exposes enough evidence but does not align bank state, retrieval state, and answer state by itself. Our contributions are:
\begin{itemize}[leftmargin=*]
\setlength{\itemsep}{0pt}
\setlength{\parsep}{-2pt}
\setlength{\parskip}{0pt}
    \item We formulate state changing long term memory as a three level problem over memory bank, retrieval, and QA, which separates failures hidden by final answer accuracy.
    \item We identify ghost memory as a state coordination failure where old, current, and transition facts remain mixed unless their roles are preserved and exposed.
    \item We introduce \ATMA, an overlay that maintains supersession structure, builds state aligned evidence packets, and conditions QA on explicit evidence state labels.
    \item We construct LTP (LoCoMo Temporal Plus) as a conflict heavy benchmark and evaluate \ATMA on LTP and LoCoMo, with host dependent gains on conflict aware and temporal queries.
\end{itemize}

\section{Related Work}
\subsection{Memory systems for LLM agents}
Long term memory has become a core component of LLM agents. Mem0 targets deployed assistants through memory extraction, consolidation, scalable retrieval, and efficient serving~\citep{chhikara2025mem0}. The agentic memory line of Xu et al. organizes experience as dynamically linked notes, so an agent can update and reuse past interactions~\citep{xu2025amem}. Other recent systems study self-updatable model memory, scalable long-term memory extensions, memory operating systems, learnable atomic memory operations, and compact online memory states~\citep{wang2024memoryllm,wang2025mplus,li2025memos,huo2026atommem,lei2026deltamem}. These systems provide useful host substrates. Their main goal is to store and retrieve helpful personal context. Our goal is narrower. We ask whether that context remains state correct after user facts change. A memory can be relevant but no longer current. \ATMA therefore treats existing memory systems as hosts and studies whether their evidence is maintained, retrieved, and interpreted according to the state requested by the query.
Earlier agent and assistant systems also motivate this setting. Generative agents use memory and reflection to produce persistent behavior, Reflexion stores verbal feedback for future decisions, Voyager builds a skill library from past experience, and MemoryBank keeps long term user memories for companion dialogue~\citep{park2023generative,shinn2023reflexion,wang2023voyager,zhong2023memorybank}. MemGPT and LongMem study context and memory management at the model or system boundary~\citep{packer2023memgpt,wang2023longmem}. These works make memory useful across sessions, but most do not isolate whether old, current, and transition facts are routed to the correct answer view.

\subsection{Temporal and Graph Based Memory}
Temporal memory systems are closest to our bank level motivation because they model how facts evolve. \Graphiti maintains temporal context graphs for agent memory and encodes changing facts with temporal structure~\citep{rasmussen2025zep,zepai2024graphiti}. \Ariadne studies lifelong memory through transition aware reasoning, disconnected evidence, and graph topology~\citep{zhu2026ariadnemem}. These systems show that temporal structure helps represent evolving user information. Yet representation alone does not guarantee a state correct answer. A graph can preserve old and new facts while retrieval selects the wrong one. The generator can also fail to decide which fact answers the query. Our work is complementary to this line. Rather than proposing another temporal substrate, \ATMA adds a level aware overlay that connects state maintenance, evidence construction, and answer resolution.
Graph retrieval systems such as HippoRAG also show that structured memory can improve integration and multi hop retrieval beyond flat dense context~\citep{jimenez2024hipporag}. Our focus is different. We do not treat graph structure as sufficient evidence of state correctness. The key variable is whether the selected packet exposes the state role needed by the query.

\subsection{Evaluation of Long Horizon Memory}
LoCoMo introduced a benchmark for very long term conversational memory, where agents answer questions from long dialogue histories~\citep{maharana2024locomo}. SeCom studies memory construction and retrieval for personalized conversational agents~\citep{pan2025secom}. Recent memory evaluations expose complementary gaps. MemTraceBench traces and attributes errors through an evolving memory pipeline~\citep{deng2026memtrace}, while MemTrace evaluates knowledge points under current, earlier, and trajectory questions~\citep{long2026memtrace}. DynamicMem stresses long horizon memory under temporal, conflicting, multi session, and noisy real world settings~\citep{xie2026dynamicmem}. MEMPROBE audits hidden user state recovery from the memory artifact, contrasting full-store access with top-k retrieved memories to separate write-side retention from read-side reachability~\citep{ma2026memprobe}. Agent-Native Memory System argues that memory evaluation should resemble deployed agents, where memory writing, updating, retrieval, and personalization interact over time~\citep{zhou2026agentnative}. Such benchmarks and systems make memory quality measurable. Final QA accuracy alone is not enough for state changing memory. A wrong answer may come from a polluted bank, misaligned retrieval, or answer side confusion over evidence that was present. A correct answer may also hide a weak bank if the downstream model guesses from partial context. \ATMA follows a diagnostic view. End results should be reported with level specific evidence: whether the bank preserves both states, whether retrieval supports the requested state view, and whether QA resolves labeled evidence correctly. Table~\ref{tab:related-memtrace-comparison} summarizes this distinction.
\begin{table*}[!t]
\centering
\caption{Comparison with long term memory evaluation and tracing work. The axes separate what is measured, which failure levels are visible, and whether the work also changes the memory pipeline.}
\label{tab:related-memtrace-comparison}
\renewcommand\tabcolsep{2.7pt}
\renewcommand\arraystretch{1.12}
\footnotesize
\resizebox{\linewidth}{!}{%
\begin{tabular}{l|>{\raggedright\arraybackslash}p{2cm}|c|c|c|>{\raggedright\arraybackslash}p{4.6cm}}
\Xhline{1.2pt}
\rowcolor{atmaheader}
\textbf{Work} & \textbf{Main Unit} & \textbf{State Change} & \textbf{Visible Failure Level} & \textbf{Pipeline Intervention} & \textbf{Relation to \ATMA} \\
\Xhline{1.2pt}
LoCoMo~\citep{maharana2024locomo} & Long conversation QA row & \nomark & Final QA & \nomark & External benchmark for long conversation memory; not designed to isolate old/current state coordination. \\
\hline
MemTraceBench~\citep{deng2026memtrace} & Memory evolution graph and operation subgraph & \nomark & Operation graph & \textsc{Prompt} & Traces operation level failures such as information loss and retrieval misalignment; it diagnoses pipeline faults and uses attribution for prompt optimization, but does not define a state role overlay. \\
\hline
MemTrace~\citep{long2026memtrace} & Knowledge point across sessions & \yesmark & Evidence use & \nomark & Probes current state, earlier state, trajectory, and evidence conditions; it shows final accuracy hides failures, while \ATMA also changes how states are stored, retrieved, and exposed to QA. \\
\hline
DynamicMem~\citep{xie2026dynamicmem} & Evolving profile item at checkpoint & \yesmark & Evidence delivery & \nomark & Evaluates retention and update over long multi-app histories; close to \ATMA's state-change setting, but does not add an explicit state-role overlay across bank, retrieval, and QA. \\
\hline
MEMPROBE~\citep{ma2026memprobe} & Hidden user-state dimension after trajectory & \nomark & Write/read recovery & \nomark & Audits recoverable user state from the memory artifact under full-store and top-k access; diagnostic like \ATMA, but not focused on old/current/transition role resolution. \\
\hline
\rowcolor{atmahlight}
LTP + \ATMA & State unit and query state view & \yesmark & Bank/Retrieval/QA & \yesmark & Targets ghost memory directly by preserving state roles in the bank, building state aligned evidence packets, and conditioning QA on explicit roles. \\
\Xhline{1.2pt}
\end{tabular}}
\end{table*}

Long context and long memory benchmarks report complementary stress tests, including broad long-context understanding, multi-session memory, temporal reasoning, updates, and abstention~\citep{bai2023longbench,wu2024longmemeval}. These evaluations show why a large context window is not equivalent to a reliable memory system. Models can still underuse evidence in long contexts or over rely on position and salience~\citep{liu2023lost}. LTP targets a narrower failure mode: state changing facts whose old and current forms must both remain available and must be routed according to the query's requested state.

\subsection{Retrieval and Evidence Conditioning}
Retrieval augmented generation builds on retrieval augmented pretraining, dense passage retrieval, and later self-reflective or note based retrieval control~\citep{guu2020realm,karpukhin2020dense,lewis2020retrieval,asai2023selfrag,yu2023chainofnote}. These methods often assume that better context leads to better answers. State changing memory weakens this assumption. Old and current facts can both be relevant, but they support different answers. The key question is not only whether retrieval finds related evidence. It must find evidence for the requested state view, and the answer model must use that view. \ATMA addresses this gap by constructing state aligned evidence packets and exposing explicit evidence labels to QA. This design separates our work from methods that improve storage, graph structure, or retrieval in isolation. It also motivates our evaluation, which reports bank, retrieval, and QA behavior as connected but distinct parts of the same system.

\section{Problem Setup}
We study long term memory under state changing user facts. A memory stream may contain records that describe an old state, a current state, or the transition between them. The target is not to delete old facts. Old facts may be correct for historical questions, while current facts should control present state questions. A useful memory system must preserve this history and answer according to the state view requested by the query.

We instantiate this setting with LTP (LoCoMo Temporal Plus), where each profile contains paired old and current states and transition records that make the update inspectable. The current LTP probes ask for either the current state or a historical state. Transition records are used as evidence and metadata, not as direct change QA targets. LTP is checked by automatic structural validation and stratified manual verification; Appendix~\ref{app:manual-verification} gives the verification protocol. This setting is related to long term conversational memory benchmarks, but LTP stresses controlled state changes rather than broad long context coverage~\citep{maharana2024locomo,wu2024longmemeval,bai2023longbench}.

The difficulty is structural. Unlike generic RAG evaluation, the distractor can be a true memory from the wrong state, not merely an irrelevant passage. This induces a three level failure decomposition. The bank must preserve coexisting states, retrieval must foreground evidence for the requested state view, and QA must resolve the answer from labeled evidence. A final answer score alone cannot distinguish these failures, so the method and experiments report bank, retrieval, and QA behavior separately.

\section{Method}
\subsection{Problem Interface and Notation}
\ATMA is a state alignment overlay for an existing memory system. The host keeps its own storage substrate, index backend, seed retriever, and answer model. This boundary matters. The method is not a new vector store and does not replace the host memory architecture. It wraps the host interface with state metadata, relation aware bank updates, and evidence organization. Given a memory stream $M=\{m_1,\ldots,m_T\}$ and a user query $q$, the host provides ordinary memory operations. It can add a memory, retrieve candidate records, and generate an answer from retrieved context. \ATMA changes what state information is attached to those records and how retrieved evidence is exposed to the answer model.

We represent each memory record as $r_i=(x_i,t_i,s_i,L_i,z_i)$. Here $x_i$ is the memory content, $t_i$ is its time or source position, $s_i$ is the state status, $L_i$ is a set of typed links to related records, and $z_i$ stores host specific metadata. The status can be active, superseded, transition, or unknown. The link set stores relations such as supersedes, superseded by, evolves from, evolves to, scoped exception, and coexistence. These fields may live in host row metadata, adapter side state, or a sidecar trace. A row based host can store status and transition fields directly. A note graph host can keep state in adapter dictionaries and logic links.

A query also has a state view. We write
\begingroup
\small
\setlength\abovedisplayskip{2pt} \setlength\belowdisplayskip{2pt}
\begin{equation}
\label{eq:query_state_view}
v_q \in \{\text{current},\text{historical},\text{transition},\text{neutral}\}.
\end{equation}
\endgroup
A current query asks for the active state. A historical query asks for a previous state. A transition query asks what changed or how two states relate. A neutral query does not specify a reliable state preference. The LTP evaluation in this paper instantiates current and historical QA targets; the transition view is used as part of the overlay interface and evidence labeling rather than as a direct LTP answer target. The method output is an answer $\hat{a}$ and an optional trace showing which records were used. The objective is not only to retrieve text relevant to $q$. The objective is to construct an evidence packet $E_q$ whose state semantics match $v_q$.

\subsection{Overview}
\begin{figure*}[t!]
	\centering
    \begin{center}
			\includegraphics[width=1\linewidth]{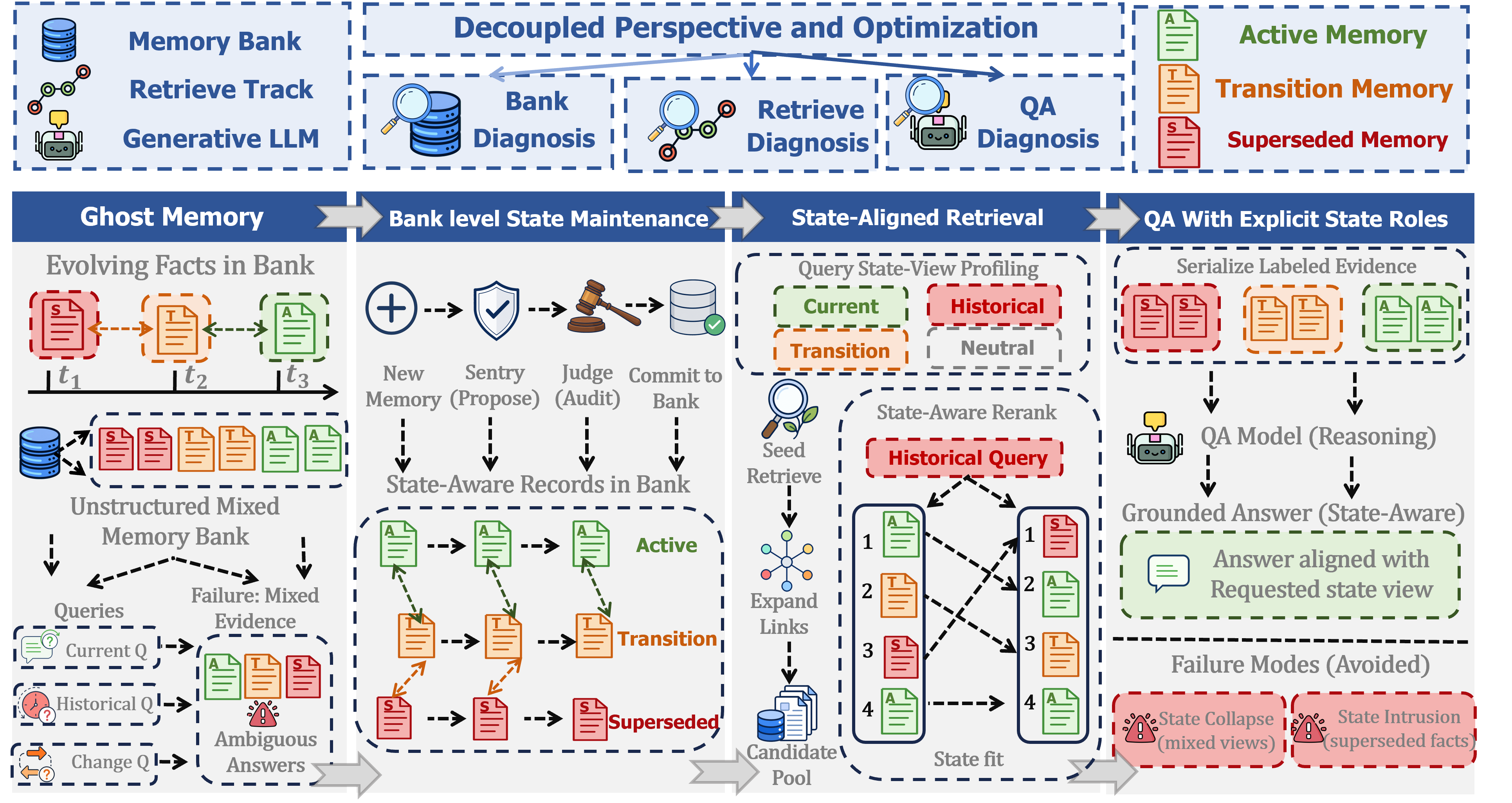}
    \end{center}
    \vspace{-10pt}
	    \caption{\small \textbf{\ATMA Framework.} Ghost memory mixes active, transition, and superseded facts across the bank, retrieval track, and generator. We separate this pipeline into bank state maintenance, state-aligned retrieval, and QA with explicit state roles, so each stage can be diagnosed and optimized before the final answer is produced.}
	    \label{fig:atmaframework}
    \vspace{-10pt}
\end{figure*}

Figure~\ref{fig:atmaframework} gives the framework view. \ATMA contains three modules. The bank module addresses ghosting inside the memory store. Old and current facts may coexist, but obsolete facts should not remain live by default. The retrieval module addresses state mismatch. The correct evidence may exist in the bank, yet the retrieved packet may match the wrong state view. The QA module addresses state collapse. The packet may include several states, but the answer model may merge them into one plausible response.

\subsection{Bank Level State Maintenance}
The bank module preserves history without letting obsolete facts act as current facts. A destructive update keeps the bank clean for current questions, but it removes evidence needed for historical recovery and transition diagnostics. A flat append only update preserves history, but stale memories can remain retrievable as if they were active. \ATMA uses a third option. It keeps both states, changes their status, and connects them through typed relations.

\noindent\textbf{State commit.} The core bank operation is a commit over a new record and any related prior records. If a new record supersedes an active record in the same state slot, \ATMA marks the old record as superseded, marks the new record as active, and adds reciprocal links from each record to the other. If the new record describes a change process rather than a direct replacement, \ATMA can attach transition metadata that records the source state, target state, and transition type. If two facts refine or qualify each other, \ATMA does not force a contradiction. It keeps both active when they can coexist or attaches a scoped exception when the new fact only applies under a condition. The commit is therefore the method's main bank side abstraction: preserve the old state, identify the live state, and leave an inspectable relation between them.

\noindent\textbf{Sentry candidate proposal.} The commit should not require scanning the whole bank. When a new memory $r_j$ arrives, \ATMA builds a small candidate set $\mathcal{P}_j$ from host retrieval and existing state links. Sentry is a lightweight proposal gate over this candidate set. It estimates whether a prior record $r_i$ discusses the same state variable and whether the two records are logically suspicious enough to need audit.

The deployed Sentry path follows the active training script. It uses a SentenceTransformer encoder and two normalized projection heads. The topic head preserves semantic slot identity, while the logic head exposes stance compatibility. For a text $x$, Sentry computes
\begingroup
\small
\setlength\abovedisplayskip{2pt} \setlength\belowdisplayskip{2pt}
\begin{equation}
\label{eq:sentry_dual_head}
\begin{gathered}
h_x=\phi_\theta(x), \\
z_x^{\mathrm{top}}=\operatorname{norm}_2\!\left(g_{\mathrm{top}}(h_x)\right),\quad
z_x^{\mathrm{log}}=\operatorname{norm}_2\!\left(g_{\mathrm{log}}(h_x)\right).
\end{gathered}
\end{equation}
\endgroup
Let $s_v(i,j)=\left\langle z_{x_i}^{v},z_{x_j}^{v}\right\rangle$. The pair metrics are then computed as
\begingroup
\small
\setlength\abovedisplayskip{2pt} \setlength\belowdisplayskip{2pt}
\begin{equation}
\label{eq:sentry_pair_score}
\begin{gathered}
a_{ij}=s_{\mathrm{top}}(i,j),\quad b_{ij}=s_{\mathrm{log}}(i,j),\quad d_{ij}=\operatorname{gap}(a_{ij},b_{ij}).
\end{gathered}
\end{equation}
\endgroup
Here $a_{ij}$ measures topic agreement, $b_{ij}$ measures logical compatibility, and $d_{ij}$ is the implementation gap score used by the detector. A candidate is routed to the Judge when it is topically close and logically suspicious:
\begingroup
\small
\setlength\abovedisplayskip{2pt} \setlength\belowdisplayskip{2pt}
\begin{equation}
\label{eq:sentry_gate_rule}
\begin{aligned}
\mathcal{A}_j = \{ i \in \mathcal{P}_j :\, & a_{ij}>\tau_{\mathrm{top}} \\
&\land \left(b_{ij}<\tau_{\mathrm{log}} \lor d_{ij}>\tau_{\mathrm{gap}}\right) \}.
\end{aligned}
\end{equation}
\endgroup

\looseness = -1
This rule proposes candidates rather than committing relations. Pure semantic similarity can over select records that mention the same entity without changing state. Pure contradiction detection can miss gradual updates. Sentry only decides which pairs need heavier audit. It does not decide the final relation.

\noindent\textbf{Sentry training objective.} The Sentry training script optimizes the two views with supervised pair labels. Each training item contains a new interaction $u_k$, a profile fact $p_k$, a topic target $y_k^{\mathrm{top}}$, and a logic target $y_k^{\mathrm{log}}$. The data loader creates three kinds of pairs. A same slot pair always has $y_k^{\mathrm{top}}=1$. If the interaction conflicts with the profile fact, $y_k^{\mathrm{log}}=0$. If it is consistent, $y_k^{\mathrm{log}}=1$. A negative fact sampled from another slot uses $y_k^{\mathrm{top}}=0$ and $y_k^{\mathrm{log}}=0$. This gives the label map
\begingroup
\small
\setlength\abovedisplayskip{2pt} \setlength\belowdisplayskip{2pt}
\begin{equation}
\label{eq:sentry_label_map}
\begin{aligned}
(y_k^{\mathrm{top}},y_k^{\mathrm{log}})=
\begin{cases}
(1,0), & c_k=\text{conflict}, \\
(1,1), & c_k=\text{consistent}, \\
(0,0), & c_k=\text{negative}.
\end{cases}
\end{aligned}
\end{equation}
\endgroup
Let $\ell(s,y)$ denote the squared target deviation used in the script. For a mini batch $\mathcal{B}$, Sentry minimizes
\begingroup
\small
\setlength\abovedisplayskip{2pt} \setlength\belowdisplayskip{2pt}
\begin{equation}
\label{eq:sentry_training_loss}
\begin{aligned}
\mathcal{L}_{\mathrm{sentry}}(\mathcal{B})=
\frac{1}{|\mathcal{B}|}\sum_{k\in\mathcal{B}}\bigg[
&\alpha\,\ell\!\left(\left\langle z_{u_k}^{\mathrm{top}},z_{p_k}^{\mathrm{top}}\right\rangle,y_k^{\mathrm{top}}\right) \\
&+\beta\,\ell\!\left(\left\langle z_{u_k}^{\mathrm{log}},z_{p_k}^{\mathrm{log}}\right\rangle,y_k^{\mathrm{log}}\right)
\bigg].
\end{aligned}
\end{equation}
\endgroup
The script splits interactions into train, validation, and test partitions by conflict type, optimizes the batch loss with AdamW, saves every epoch, and selects the deployed checkpoint by validation loss:
\begingroup
\small
\setlength\abovedisplayskip{2pt} \setlength\belowdisplayskip{2pt}
\begin{equation}
\label{eq:sentry_checkpoint_selection}
\begin{gathered}
\Theta^{\star}_{\mathrm{sentry}}=\arg\min_{\Theta_e}\,\mathcal{L}_{\mathrm{sentry}}^{\mathrm{val}}(\Theta_e), \\
\Theta_e=\{\theta,W_1^{\mathrm{top}},W_2^{\mathrm{top}},W_1^{\mathrm{log}},W_2^{\mathrm{log}}\}.
\end{gathered}
\end{equation}
\endgroup

\noindent\textbf{Judge audit and commit.} The selected Sentry checkpoint becomes the lightweight proposal gate in the bank update path. The heavier Judge is trained separately to produce structured audit decisions through supervised warm up and reward based optimization for verdict format, correctness, old state preservation, and transition summaries. For each suspicious pair in $\mathcal{A}_j$, the Judge receives the old record, the new input, and Sentry metrics. It returns a conflict status, conflict type, fusion action, optional clarification needs, and suggested links.

If the new record supersedes an active record in the same slot, the commit can be written as
\begingroup
\small
\setlength\abovedisplayskip{2pt} \setlength\belowdisplayskip{2pt}
\begin{equation}
\label{eq:bank_supersession_commit}
\begin{aligned}
s_i &\leftarrow \text{superseded}, \\
s_j &\leftarrow \text{active}, \\
L_i &\leftarrow L_i \cup \{(\text{superseded by},j)\}, \\
L_j &\leftarrow L_j \cup \{(\text{supersedes},i)\}.
\end{aligned}
\end{equation}
\endgroup
For records that refine, coexist with, or qualify a prior fact, \ATMA does not force a supersession relation. It keeps the old record active when the states can coexist, or attaches a scoped exception when the new fact only applies under a condition. Appendix~\ref{app:implementation-details} reports deployed backbone, thresholds, and audit model settings; the main mechanism remains the Sentry proposal, Judge audit, and state commit path above.

\looseness = -1
The bank layer follows simple invariants. A single valued state slot should not have two active current states unless there is an explicit coexistence relation. A superseded state should remain recoverable for historical queries. A transition link should preserve how the current state evolved from the prior state when that evolution matters for interpretation. These invariants make the bank inspectable. They do not solve the full task. A clean bank can still lead to a wrong answer if retrieval selects the wrong state.

\subsection{Retrieve Level Evidence Construction}
\vspace{-5pt}

The retrieval module addresses the second failure mode. A bank can preserve the correct state while the retriever exposes the wrong one. In evolving memory, relevance and state fit are not the same signal. A current query and a historical query may use almost identical words, but they ask the system to foreground different records. \ATMA treats retrieval as evidence packet construction. It first lets the host memory system produce semantic seed rows. It then expands the pool through state links and chooses an ordered packet whose state view matches the query.

\noindent\textbf{Query state view.} The implementation uses a lightweight rule based query profiler before ranking, not a classifier or an LLM. It assigns a view $v_q$ from current, historical, transition, and neutral by counting temporal and change hints, and records guards for negation, scope, and update wording. We denote the resulting profile by $\psi(q)$. The guards are not a separate answer model. They tell the packet builder when old records, successor records, or contrast pairs are likely to be useful. A query asking what was true before an update should not suppress a superseded record only because a newer successor exists.

\looseness = -1
\noindent\textbf{Seed and hop pool.} The host retriever provides the seed set $A_q$. \ATMA preserves this order because it is the host system's strongest semantic signal. It then constructs two auxiliary sets. Relation hops $R_q$ come from state links such as supersedes, superseded by, and transition relations. Semantic hops $N_q$ come from weaker note or neighbor links and are used only as support. The expansion is view aware:
\begingroup
\small
\setlength\abovedisplayskip{2pt} \setlength\belowdisplayskip{2pt}
\begin{equation}
\label{eq:candidate_pool}
\begin{gathered}
A_q=R_H(q,B_t),\quad R_q=\Gamma_{\mathrm{rel}}(A_q,B_t,v_q),\quad N_q=\Gamma_{\mathrm{sem}}(A_q,B_t), \\
\mathcal{C}_q=\operatorname{Dedup}\!\left(A_q \oplus R_q \oplus N_q\right).
\end{gathered}
\end{equation}
\endgroup
Here $\oplus$ denotes ordered concatenation before duplicate removal. When the query is not historical, successors of stale seeds can be added because they may be the live state. When the query is not current, predecessors can be added because they may answer a historical query or support transition diagnostics. This trace is part of the method, not only logging. If the target record is absent from $\mathcal{C}_q$, the error belongs to candidate construction. If it is present but ordered behind the wrong state, the error belongs to ranking. If it is present and foregrounded but the answer is wrong, the error belongs to QA resolution.

\noindent\textbf{Stable pre rank.} The default rank is conservative. Older heuristic state reranking rules were removed from the active path because large hand written bonuses were brittle across query views and host systems. The stable pre rank preserves the bucket order used to form $\mathcal{C}_q$:
\begingroup
\small
\setlength\abovedisplayskip{2pt} \setlength\belowdisplayskip{2pt}
\begin{equation}
\label{eq:stable_prerank}
\pi_0=\operatorname{rank}\!\left(\operatorname{Dedup}(A_q \oplus R_q \oplus N_q)\right),\quad
E_q^{0}=\operatorname{TopK}(\pi_0,k).
\end{equation}
\endgroup
This rank is the fallback packet builder. It gives direct semantic hits priority, keeps relation rows available, and avoids hiding host behavior behind uninspectable score shaping. The pre rank remains a stable reference point for replay and ablation.

\noindent\textbf{Bounded controller rank.} The optional retrieve controller operates only on $\mathcal{C}_q$. It receives a compact JSON view containing the query profile, candidate id, pre rank, source bucket, status, relation, transition type, supersession fields, and truncated content. It must not answer the query or use outside knowledge. Its output is a query mode, confidence, fallback flag, selected candidate ids, role labels, bounded fit scores, and packing hints. Formally, the controller returns
\begingroup
\small
\setlength\abovedisplayskip{2pt} \setlength\belowdisplayskip{2pt}
\begin{equation}
\label{eq:retrieve_controller_output}
\begin{gathered}
\Omega_q=F_\eta(q,\psi(q),\mathcal{C}_q), \\
M\leq\min(k,K_{\mathrm{sel}}),\qquad i_m\in\mathcal{C}_q, \\
\Omega_q=\left(\hat{v}_q,c_q,f_q,\{(i_m,o_m,\ell_m)\}_{m=1}^{M},h_q\right).
\end{gathered}
\end{equation}
\endgroup
Here $\hat{v}_q$ is the controller's view prediction, $c_q$ is confidence, $f_q$ is the fallback recommendation, $i_m$ is a selected candidate id, $o_m$ is its controller order, $\ell_m$ contains role and fit labels, and $h_q$ contains packing hints. The constraint $i_m\in\mathcal{C}_q$ is the key boundary. The controller can reorder existing evidence, but it cannot retrieve missing evidence or invent a memory.

The final ordering is applied only when the controller is enabled online, its JSON parses, and it selects valid candidates. Otherwise, the method returns to the stable pre rank:
\begingroup
\small
\setlength\abovedisplayskip{2pt} \setlength\belowdisplayskip{2pt}
\begin{equation}
\label{eq:evidence_packet_builder}
\begin{aligned}
\pi_c &= \operatorname{Sort}_{o}\!\left(\{i_m\}_{m=1}^{M}\right) \oplus \operatorname{Remain}(\pi_0,\{i_m\}_{m=1}^{M}), \\
\pi_q =
\begin{cases}
\pi_c, & \operatorname{online}(\Omega_q)\land\operatorname{valid}(\Omega_q)\land M>0, \\
\pi_0, & \text{otherwise},
\end{cases} \\
E_q &= \operatorname{TopK}(\pi_q,k).
\end{aligned}
\end{equation}
\endgroup
This makes the rank mechanism bounded and auditable. The controller can foreground a current record, keep an old record as contrast, or select both sides of a transition. Any unselected portion of the packet is still filled by the pre rank. A retrieval trace records the query profile, candidate pool, pre rank, controller selections, parse status, and final order. Retrieve level evaluation can then measure target hit, state view success, stale leakage, and future state intrusion without conflating them with answer wording.

\subsection{QA Level Evidence State Conditioning}
\looseness = -1
The QA module addresses the third failure mode, answer side state collapse. At this point, the needed record may already be present in $E_q$. The answer model still sees live facts, old facts, and transition evidence as plain text. A relevance only context window makes the model infer state from wording alone. This is fragile because current and historical records often share the same entity, slot, and surface terms. \ATMA therefore does not pass raw evidence directly to the generator. It first exposes the state carried by the retrieval trace, then asks the answer model to resolve the state requested by the query.

\noindent\textbf{Trace to label conversion.} Each retrieved item keeps the metadata produced by the bank and retrieval levels: status, relation, transition type, source, timestamp, and supersession fields. The QA serializer maps these fields into a small label vocabulary:
\begingroup
\small
\setlength\abovedisplayskip{2pt} \setlength\belowdisplayskip{2pt}
\begin{equation}
\label{eq:qa_label_space}
\begin{gathered}
\lambda_i=\Lambda(s_i,\rho_i,\delta_i,\mu_i), \\
\lambda_i\in\{\text{cur},\text{hist},\text{tran},\text{link},\text{raw}\}.
\end{gathered}
\end{equation}
\endgroup
Here $s_i$ is status, $\rho_i$ is relation, $\delta_i$ is transition metadata, and $\mu_i$ is source metadata. The priority rule is simple. A transition type or transition row becomes transition memory. Evolution links become transition linked memory. Superseded, archived, stale, or inactive rows become historical memory. Active rows, successor rows, and rows that supersede another record become current memory. The fallback label is retrieved memory. This mapping is not a learned answer policy. It is a deterministic projection of trace fields that earlier levels already produced.

\noindent\textbf{State explicit evidence prompt.} After labeling, the serializer keeps the content order selected by retrieval but annotates each line with its state role and optional time field:
\begingroup
\small
\setlength\abovedisplayskip{2pt} \setlength\belowdisplayskip{2pt}
\begin{equation}
\label{eq:qa_state_serialization}
\widetilde{E}_q=\left[(i,\lambda_i,t_i,\delta_i,x_i)\right]_{i=1}^{|E_q|},\quad
\mathcal{P}_q=\operatorname{Prompt}(q,v_q,\widetilde{E}_q,\mathcal{R}_{\mathrm{qa}}).
\end{equation}
\endgroup
The rule set $\mathcal{R}_{\mathrm{qa}}$ constrains the generator to use retrieved memories only. In the full evaluation path, QA uses \texttt{qwen2.5:3b} with retrieval budget $k=5$; the separate \texttt{llama3.3:70b} model is used only as an offline evaluator. If current and historical memories both appear, the answer must follow the state asked for by the question. Current memory is treated as the active state unless the question explicitly asks about the past. Historical memory is treated as old or superseded state. Transition memory and transition linked memory are used to explain how a state changed. These rules make hidden state semantics visible to the answer model. The model no longer has to infer them from raw prose alone. The final answer step is
\begingroup
\small
\setlength\abovedisplayskip{2pt} \setlength\belowdisplayskip{2pt}
\begin{equation}
\label{eq:qa_state_conditioned_answer}
\hat{a}=G_H(\mathcal{P}_q),\quad
\hat{a}\preceq \widetilde{E}_q,\qquad \operatorname{view}(\hat{a})\simeq v_q.
\end{equation}
\endgroup
The notation $\hat{a}\preceq \widetilde{E}_q$ means that the answer should be grounded in the serialized evidence. The notation $\operatorname{view}(\hat{a})\simeq v_q$ means that the answer should match the requested state view. In hosts whose native answerer does not consume evidence trace, the \ATMA adapter routes generation through this common state conditioned path.

This boundary matters for both method and evaluation. QA level conditioning can prevent the generator from choosing an old state when a current state is labeled. It can also prevent the generator from overwriting a requested historical state with its active successor. It cannot recover a missing target record. It cannot repair a packet whose rank has excluded the needed state. Those failures belong to bank maintenance or retrieval construction. The role of the QA level is narrower and testable: given an evidence packet with state trace, it turns implicit memory status into an explicit answer constraint. The complete procedural view is provided in \hyperref[alg:atma_inference]{Appendix Algorithm~\ref*{alg:atma_inference}}.

\subsection{Discussion and Boundaries}
The three modules are connected but not interchangeable. A bank only method can preserve old and new facts while still retrieving the wrong one. A retrieval only method can surface useful evidence while leaving the generator to infer state semantics implicitly. A QA only method can interpret labels but cannot repair evidence that was not retrieved. This is also where \ATMA differs from temporal graph memory. A temporal bank can encode evolving facts, but state correct QA still depends on whether retrieval foregrounds the requested state and whether the answer model follows that state. \ATMA therefore defines where state is stored, how it is selected, and how it is resolved.

The same decomposition defines the method's scope. \ATMA targets cases where useful evidence exists but remains mixed across bank, retrieval, and answer time. It does not recover target records that the host never stores or retrieves. LTP exposes this chain through bank state roles, evidence support, and state resolved QA metrics. LoCoMo provides an external long conversation check, but its lexical metrics do not test state resolution in the same controlled way.

\section{Experiments}

\subsection{Experimental Setup}
\noindent \textbf{Datasets.}
We evaluate on two complete long term memory benchmarks. LTP contains 10 user profiles and 800 QA probes over stable facts, obsolete facts, and transition records. Its direct QA targets are current and historical states. We also report full LoCoMo results over all 10 conversation samples and 1,986 QA pairs as an external generalization check.

\noindent \textbf{Baselines.}
We compare against host memory systems and temporal memory baselines that completed the relevant full evaluation. The LTP table includes Mem0~\citep{chhikara2025mem0}, InsideOut~\citep{zhao2026insideout}, AtomMem~\citep{huo2026atommem}, \Ariadne~\citep{zhu2026ariadnemem}, \Graphiti~\citep{rasmussen2025zep,zepai2024graphiti}, SeCom~\citep{pan2025secom}, A-Mem~\citep{xu2025amem}, MemoryLLM~\citep{wang2024memoryllm}, M+~\citep{wang2025mplus}, and completed \ATMA variants. Numeric LoCoMo rows use complete 10 sample results. Rows without full runs are left blank rather than mixing subset results into the main comparison.

\noindent \textbf{ATMA Details.}
\ATMA is evaluated as a query state alignment overlay. It preserves bank state labels, builds evidence packets for the requested state view, and exposes current, historical, and transition evidence to the answer model. Unless otherwise stated, QA uses \texttt{qwen2.5:3b}, retrieval uses top $k=5$, and the answer judge is a separate \texttt{llama3.3:70b} service. For LTP, we report \targacc, conflict and fact splits, average judge score, \evidsupport, and \bankrole. For LoCoMo, we report F1, BLEU-1~\citep{papineni2002bleu}, and LoCoMo-specific judged QA accuracy where the full judged run is complete. Appendix~\ref{app:experimental-details} gives the evaluation details.

\noindent \textbf{Evaluation Protocol.}
We report LTP at three levels. \bankrole checks whether the final bank exposes a method native state role signal for the matched current and old values. The signal can be an explicit status, a supersession or transition link, or a usable temporal order depending on the host interface. \evidsupport checks whether retrieved evidence supports the requested state view. \targacc, conflict and fact splits, and judge accuracy measure final state resolution.

\subsection{Main Results on Full LTP}
\begin{table*}[!t]
\centering
\caption{Full LTP evaluation over all 10 profiles (800 judged probes per method) under the 70B judge. The first three columns mark whether each method explicitly addresses state-aware bank maintenance, state-aligned retrieval, and answer-side state resolution. \bankrole is scored with method-native status or state-link signals; timestamp-only ordering is not credited. Metric columns are grouped by evaluation level. Rank is the average rank across the seven reported metrics, with lower better. For each completed +\ATMA row, colored deltas show the absolute change from its base method.}
\label{tab:main-ltp-full}
\renewcommand\tabcolsep{1.85pt}
\renewcommand\arraystretch{1.12}
\footnotesize
\resizebox{\linewidth}{!}{%
\begin{tabular}{l|ccc|c|rr|r|rrrr|r}
\Xhline{1.2pt}
\rowcolor{atmaheader}
& \multicolumn{3}{c|}{\textbf{Modeled Levels}} & & \multicolumn{2}{c|}{\textbf{Bank Level}} & \multicolumn{1}{c|}{\textbf{Retrieve Level}} & \multicolumn{4}{c|}{\textbf{QA Level}} & \textbf{Avg.} \\
\rowcolor{atmaheader}
\multirow{-2}{*}{\textbf{Method}} & \textbf{Bank} & \textbf{Ret.} & \textbf{QA} & \multirow{-2}{*}{\textbf{N}} & \textbf{\bankcoexist} & \textbf{\bankrole} & \textbf{\evidsupport} & \textbf{QA Acc.} & \textbf{Conflict} & \textbf{Fact} & \textbf{Judge Acc.} & \textbf{Rank} \\
\Xhline{1.2pt}
Mem0 & \nomark & \nomark & \nomark & 800 & 0.007 & 0.000 & 0.302 & 0.188 & 0.258 & 0.117 & 0.174 & 10.64 \\
AtomMem & \nomark & \nomark & \nomark & 800 & \textbf{1.000} & 0.000 & 0.539 & 0.381 & 0.522 & 0.240 & 0.311 & 8.14 \\
SeCom & \nomark & \nomark & \nomark & 800 & \underline{0.990} & 0.000 & 0.864 & 0.655 & 0.733 & 0.578 & 0.564 & 5.07 \\
MemoryLLM & \nomark & \nomark & \nomark & 800 & \textbf{1.000} & 0.000 & 0.511 & 0.400 & 0.487 & 0.312 & 0.339 & 8.43 \\
M+ & \nomark & \nomark & \nomark & 800 & \textbf{1.000} & 0.000 & 0.527 & 0.409 & 0.490 & 0.328 & 0.347 & 7.71 \\
\Ariadne & \yesmark & \yesmark & \nomark & 800 & \textbf{1.000} & 0.000 & 0.656 & 0.453 & 0.512 & 0.393 & 0.381 & 6.86 \\
\hline
A-Mem & \yesmark & \yesmark & \nomark & 800 & \textbf{1.000} & 0.568 & \underline{0.916} & \underline{0.787} & \underline{0.812} & \underline{0.762} & \textbf{0.700} & \underline{2.36} \\
\rowcolor{atmahlight} A-Mem+\ATMA & \yesmark & \yesmark & \yesmark & 800 & \textbf{1.000}\updelta{0.000} & \textbf{0.825}\updelta{0.257} & \textbf{0.925}\updelta{0.009} & \textbf{0.818}\updelta{0.031} & \textbf{0.860}\updelta{0.048} & \textbf{0.775}\updelta{0.013} & \underline{0.698}\downdelta{0.002} & \textbf{1.79} \\
\Graphiti & \yesmark & \yesmark & \nomark & 800 & \textbf{1.000} & 0.000 & 0.711 & 0.524 & 0.480 & 0.568 & 0.473 & 6.57 \\
\rowcolor{atmahlight} \Graphiti+\ATMA & \yesmark & \yesmark & \yesmark & 800 & \textbf{1.000}\updelta{0.000} & \underline{0.775}\updelta{0.775} & 0.809\updelta{0.098} & 0.635\updelta{0.111} & 0.720\updelta{0.240} & 0.550\downdelta{0.018} & 0.559\updelta{0.086} & 5.43 \\
InsideOut & \yesmark & \nomark & \nomark & 800 & \textbf{1.000} & 0.000 & 0.239 & 0.117 & 0.175 & 0.060 & 0.115 & 10.43 \\
\rowcolor{atmahlight} InsideOut+\ATMA & \yesmark & \yesmark & \yesmark & 800 & \textbf{1.000}\updelta{0.000} & 0.000\updelta{0.000} & 0.848\updelta{0.609} & 0.662\updelta{0.545} & 0.680\updelta{0.505} & 0.645\updelta{0.585} & 0.561\updelta{0.446} & 4.57 \\
\Xhline{1.2pt}
\end{tabular}}
\end{table*}

Table~\ref{tab:main-ltp-full} reports the full LTP results. \bankrole is personalized to each bank interface: temporal and graph based methods can receive credit when their stored records expose a usable ordering or link between the old and current values, while systems without a matched role signal receive no credit. This makes the bank result diagnostic rather than a direct proxy for final QA quality. Several methods expose state-role cues in the bank, but only some retrieve and use them correctly. A-Mem+\ATMA reaches the strongest aggregate \targacc at 0.818 over 800 probes and improves the conflict split from 0.812 to 0.860, while A-Mem remains slightly higher on judge score.

Larger gains appear on hosts with weaker state view alignment. InsideOut+\ATMA improves \targacc from 0.117 to 0.662. This large gain is expected under LTP because InsideOut compresses dialogue into an evolving user profile, which can weaken access to old states once a slot has changed. \ATMA repairs this chain at all three levels: it keeps state roles in the bank representation, constructs retrieval packets that foreground the requested current or historical state, and exposes those roles to the answer model. \Graphiti+\ATMA improves \targacc from 0.524 to 0.635 and conflict accuracy from 0.480 to 0.720. These results support the overlay framing: \ATMA helps when useful evidence exists but the host does not consistently preserve state roles, foreground the requested state during retrieval, or expose those roles to QA.

\subsection{Full LoCoMo Generalization}
\begin{table*}[!t]
\centering
\caption{Full LoCoMo evaluation on single hop, temporal, and open domain questions. Metrics are F1, BLEU-1, and judged QA accuracy; colored deltas compare +\ATMA with base rows.}
\label{tab:locomo-full}
\renewcommand\tabcolsep{1.65pt}
\renewcommand\arraystretch{1.12}
\footnotesize
\resizebox{\linewidth}{!}{%
\begin{tabular}{l|ccc|ccc|ccc}
\Xhline{1.2pt}
\rowcolor{atmaheader}
& \multicolumn{3}{c|}{\textbf{Single hop}} & \multicolumn{3}{c|}{\textbf{Temporal}} & \multicolumn{3}{c}{\textbf{Open domain}} \\
\rowcolor{atmaheader}
\multirow{-2}{*}{\textbf{Method}} & \textbf{F1} & \textbf{BLEU-1} & \textbf{QA Acc.} & \textbf{F1} & \textbf{BLEU-1} & \textbf{QA Acc.} & \textbf{F1} & \textbf{BLEU-1} & \textbf{QA Acc.} \\
\Xhline{1.2pt}
Mem0 & 0.0462 & 0.0345 & \textbf{0.5638} & 0.0173 & 0.0111 & \textbf{0.7445} & 0.0686 & 0.0529 & \underline{0.4271} \\
AtomMem & 0.0972 & 0.0332 & 0.1454 & 0.0167 & 0.0093 & 0.0561 & 0.0637 & 0.0421 & 0.1979 \\
SeCom & \underline{0.1325} & 0.1039 & 0.1773 & 0.0536 & 0.0388 & 0.1184 & 0.0680 & 0.0486 & 0.2083 \\
MemoryLLM & 0.0973 & 0.0753 & 0.1206 & \underline{0.1530} & \underline{0.1157} & 0.1495 & 0.0953 & 0.0684 & 0.2292 \\
M+ & 0.1009 & 0.0812 & 0.1241 & 0.1527 & 0.1148 & 0.1682 & \underline{0.0963} & 0.0699 & 0.1979 \\
\Ariadne & \textbf{0.2663} & \textbf{0.2227} & 0.1667 & 0.1242 & 0.1047 & 0.3240 & \textbf{0.2004} & \textbf{0.1726} & 0.1979 \\
\hline
A-Mem & 0.1176 & 0.1004 & \underline{0.4314} & 0.1429 & 0.1037 & 0.1967 & 0.0229 & 0.0081 & 0.3333 \\
\rowcolor{atmahlight} A-Mem+\ATMA & 0.1191\updelta{0.002} & \underline{0.1214}\updelta{0.021} & \underline{0.4314}\updelta{0.000} & 0.1493\updelta{0.006} & 0.1047\updelta{0.001} & 0.3279\updelta{0.131} & 0.0385\updelta{0.016} & 0.0207\updelta{0.013} & \textbf{0.4286}\updelta{0.095} \\
\Graphiti & 0.1051 & 0.0888 & 0.2482 & 0.0295 & 0.0200 & \underline{0.3707} & 0.0941 & \underline{0.0757} & 0.3333 \\
\rowcolor{atmahlight} \Graphiti+\ATMA & 0.1060\updelta{0.001} & 0.0869\downdelta{0.002} & 0.3440 & \textbf{0.1705}\updelta{0.141} & \textbf{0.1268}\updelta{0.107} & 0.3178 & 0.0906\downdelta{0.004} & 0.0646\downdelta{0.011} & 0.3958 \\
InsideOut & 0.0737 & 0.0548 & 0.3404 & 0.0117 & 0.0089 & 0.2212 & 0.0902 & 0.0725 & 0.3229 \\
\rowcolor{atmahlight} InsideOut+\ATMA & 0.1158\updelta{0.042} & 0.0831\updelta{0.028} & 0.1738 & 0.0510\updelta{0.039} & 0.0356\updelta{0.027} & 0.1900 & 0.0951\updelta{0.005} & 0.0720\downdelta{0.001} & 0.2708 \\
\Xhline{1.2pt}
\end{tabular}}
\end{table*}

\begin{table*}[!t]
\centering
\caption{Full LoCoMo evaluation continued, with multi hop, adversarial, and average scores. Rank averages the three average metrics, with lower better.}
\label{tab:locomo-full-continued}
\setlength{\tabcolsep}{1.65pt}
\renewcommand\arraystretch{1.12}
\footnotesize
\resizebox{\linewidth}{!}{%
\begin{tabular}{l|ccc|ccc|cccc}
\Xhline{1.2pt}
\rowcolor{atmaheader}
& \multicolumn{3}{c|}{\textbf{Multi hop}} & \multicolumn{3}{c|}{\textbf{Adversarial}} & \multicolumn{4}{c}{\textbf{Average}} \\
\rowcolor{atmaheader}
\multirow{-2}{*}{\textbf{Method}} & \textbf{F1} & \textbf{BLEU-1} & \textbf{Judge Acc.} & \textbf{F1} & \textbf{BLEU-1} & \textbf{Judge Acc.} & \textbf{F1} & \textbf{BLEU-1} & \textbf{Judge Acc.} & \textbf{Rank} \\
\Xhline{1.2pt}
Mem0 & 0.0375 & 0.0264 & 0.5065 & 0.0045 & 0.0045 & \textbf{0.5919} & 0.0296 & 0.0214 & \textbf{0.5685} & 8.33 \\
AtomMem & 0.0887 & 0.0513 & 0.4257 & \underline{0.0448} & \underline{0.0448} & 0.3924 & 0.0672 & 0.0401 & 0.3077 & 10.33 \\
SeCom & \underline{0.2809} & \underline{0.2047} & \textbf{0.6266} & 0.0112 & 0.0112 & 0.3700 & 0.1523 & 0.1126 & 0.4028 & 3.67 \\
MemoryLLM & 0.1977 & 0.1461 & 0.4090 & 0.0336 & 0.0336 & 0.3946 & 0.1344 & 0.1021 & 0.3142 & 7.67 \\
M+ & 0.1987 & 0.1464 & 0.4102 & 0.0314 & 0.0314 & 0.4260 & 0.1349 & 0.1025 & 0.3238 & 6.67 \\
\Ariadne & \textbf{0.3540} & \textbf{0.3090} & 0.3389 & 0.0291 & 0.0291 & 0.1570 & \textbf{0.2240} & \textbf{0.1943} & 0.2644 & 4.67 \\
\hline
A-Mem & 0.2179 & 0.1592 & 0.5858 & 0.0247 & 0.0247 & 0.4938 & 0.1411 & 0.1058 & 0.4700 & 4.33 \\
\rowcolor{atmahlight} A-Mem+\ATMA & 0.2136\downdelta{0.004} & 0.1510\downdelta{0.008} & 0.5740\downdelta{0.012} & \textbf{0.0494}\updelta{0.025} & \textbf{0.0494}\updelta{0.025} & \underline{0.5432}\updelta{0.049} & 0.1464\updelta{0.005} & 0.1110\updelta{0.005} & \underline{0.5013}\updelta{0.031} & \underline{3.33} \\
\Graphiti & 0.1231 & 0.0951 & 0.3924 & 0.0202 & 0.0202 & 0.4596 & 0.0809 & 0.0643 & 0.3807 & 8.00 \\
\rowcolor{atmahlight} \Graphiti+\ATMA & 0.2398\updelta{0.117} & 0.1801\updelta{0.085} & \underline{0.5874} & 0.0314\updelta{0.011} & 0.0314\updelta{0.011} & 0.4193 & \underline{0.1556}\updelta{0.075} & \underline{0.1193}\updelta{0.055} & 0.4622 & \textbf{2.67} \\
InsideOut & 0.0554 & 0.0381 & 0.3187 & 0.0135 & 0.0135 & 0.3498 & 0.0432 & 0.0319 & 0.3132 & 10.67 \\
\rowcolor{atmahlight} InsideOut+\ATMA & 0.2283\updelta{0.173} & 0.1558\updelta{0.118} & 0.5172 & 0.0359\updelta{0.022} & 0.0359\updelta{0.022} & 0.3049 & 0.1340\updelta{0.091} & 0.0951\updelta{0.063} & 0.3560 & 7.67 \\
\Xhline{1.2pt}
\end{tabular}}
\end{table*}

Tables~\ref{tab:locomo-full} and~\ref{tab:locomo-full-continued} report only full LoCoMo rows across single hop, temporal, open domain, multi hop, adversarial, and average scores. \Ariadne has the strongest average F1 and BLEU-1. \ATMA effects are host and metric dependent: \Graphiti+\ATMA raises temporal F1 from 0.0295 to 0.1705 and average F1 from 0.0809 to 0.1556, while A-Mem+\ATMA remains close to A-Mem but is slightly lower on average F1 and BLEU-1. We therefore use LoCoMo as a mixed generalization check, not as evidence of uniform improvement. Appendix~\ref{app:level-ablation-profiles000-002} gives a three profile diagnostic ablation alongside the full benchmark results in Tables~\ref{tab:main-ltp-full}--\ref{tab:locomo-full-continued}.

\section{Conclusion}
We presented \ATMA, a tri level state aware framework for long term agent memory under changing user states. \ATMA separates memory bank maintenance, state aware retrieval, and answer time evidence resolution. This separation makes ghost memory measurable rather than treating it as a single end to end QA failure.

The full LTP results support a bounded conclusion. \ATMA preserves strong A-Mem performance and improves its conflict split. It gives larger gains for hosts whose retrieval and answer layers are less state aligned. Full LoCoMo results are mixed by host and metric, so they support cautious generalization rather than a universal improvement claim. Future agent memory evaluations should report not only whether an answer is correct, but also which memory layer made the answer possible or caused it to fail.

\bibliographystyle{unsrtnat}
\bibliography{refs_arxiv}

\appendix
\section{Experimental Details}
\label{app:experimental-details}

\subsection{Benchmark Protocol}
\label{app:benchmark-protocol}

All main LTP rows use the full 10 profile evaluation with 800 judged probes per method. LTP probes are written to test state changing memory. The release contains 400 historical fact recall probes and 400 current state conflict probes. A method may see a current fact, an obsolete fact, and a transition record about the same user attribute, but the direct QA target is the requested current or historical state rather than a free form change description. This design makes it possible to separate bank state role assignment, retrieval evidence, and final QA behavior rather than treating every failure as a single answer error.

LoCoMo is used as an external long conversation benchmark. The full LoCoMo table reports all 10 conversation samples and 1,986 QA pairs. The question categories follow the original benchmark grouping: single hop, temporal, open domain, multi hop, and adversarial. We report full rows only. Rows without a completed full run are left blank rather than filled with subset runs, partial judge passes, or skip judge outputs.

\subsection{LTP Construction Process}
\label{app:ltp-construction}

LTP is constructed from LoCoMo anchored user profiles and then converted into a controlled state changing memory benchmark. The construction begins by binding each of 10 profiles to one LoCoMo conversation. For every profile, we build 40 mutable slot instances over job, location, relationship, pets, hobbies, likes or dislikes, and numerical facts. Each slot keeps the source LoCoMo conversation id, a profile id, a slot key, a topic key, and the preferred update type. This gives the benchmark a long conversation seed while keeping the state changing facts under explicit control.

For each mutable slot, the pipeline builds a two cue state unit. The first cue records the old state and the second cue records the updated or conflicting current state. The update type is category dependent: job and location use temporal updates, relationship and pets use explicit negation, numerical facts use value discrepancy, and hobbies or preferences use scope mismatch. The release builder then attaches two QA views to each unit. The fact recall view targets the old state before the update, while the conflict aware view targets the current state after the update. Thus LTP asks for current or historical state resolution, not a free form description of the change itself.

The final profile level inputs pack all 40 units for a profile into its paired LoCoMo conversation and replay them into one final bank. Cue1 events are spread through the early to middle part of the conversation, cue2 events are inserted later, and the original LoCoMo tail is retained. The released full setting contains 10 profiles, 400 state units, 800 cue events, and 800 evaluation probes. Each profile has 40 state units and 80 probes, so final bank evaluation tests whether a method can resolve many coexisting old and current states inside the same user memory.

Before release, LTP is filtered by automatic structural checks, model assisted preaudit, targeted repair, and manual review. The structural validators check profile count, unique sample and eval ids, per profile coverage, the two QA rows per state unit, gold old and current answer consistency, cue1 before cue2 timing, target state labels, and leakage rules for the two evaluation views. The preaudit stage sends high risk samples to a secondary review queue and checks whether cue1 is a clear old state, cue2 is a valid current state, the pair is a same slot update, both questions target the intended state, the gold answers are stable, and the packed profile wording remains unambiguous. We then repair or remove flagged samples and run the stratified manual verification described below.

\subsection{LTP Replay and Insertion Setting}
\label{app:ltp-replay-setting}

LTP is used as a final bank evaluation rather than a per question context injection task. For each profile, we first sort the original LoCoMo sessions by their timestamps. We then interleave the 40 old state cues and the 40 current or conflicting cues into the same timeline. The old state cues are placed across the early to middle portion of the conversation, from the 0.12 to 0.42 time quantiles. The current or conflicting cues are placed later, from the 0.58 to 0.88 time quantiles, with an additional offset so that the update cue is always after its paired old cue. The original tail of the LoCoMo conversation remains after the inserted updates, so the final bank contains old facts, later updates, unrelated long conversation context, and post update context together.

Each method receives the same chronological replay stream and builds one final memory bank for the profile. No additional memory writing is performed during QA. The adapter only converts the replay stream into the host system's native input format: turn based systems receive text turns such as ``Speaker Caroline says : ...'', Mem0 style systems receive short message batches under speaker specific user views, SeCom receives session exchanges, and InsideOut or PersonaTree style systems receive dialogue events. After this replay, the 80 profile probes are answered from the final bank with a fixed retrieval budget. The fact recall probe for a state unit asks for the cue1 old state, while the conflict aware probe asks for the cue2 current state. This setting tests whether the memory system can keep both states available in one persistent bank and select the state requested by the query.

\subsection{Stratified Manual Verification}
\label{app:manual-verification}

We conduct a stratified manual verification on an 80 unit LTP sample. The sample is drawn by update type from the 400 state units, with 4 temporal updates, 10 explicit negations, 26 value discrepancies, and 40 scope mismatches, matching the skew of the full release. Each sampled unit contains the old state cue, current state cue, fact recall question, and current state question. Two annotators independently verify the state slot, old value, current value, update type, old state gold answer, current state gold answer, and a binary validity decision.

\begin{center}
\small
\begin{tabular}{@{}lll@{}}
\toprule
\textbf{Field} & \textbf{Agreement criterion} & \textbf{IAA} \\
\midrule
State slot & Canonical exact match & 100.0\% \\
Old value & Semantic normalized match & 100.0\% \\
Current value & Semantic normalized match & 100.0\% \\
Update type & Canonical exact match & 100.0\% \\
Old state gold answer & Semantic normalized match & 100.0\% \\
Current state gold answer & Semantic normalized match & 97.5\% \\
Validity & Exact match & 97.5\% \\
\bottomrule
\end{tabular}
\end{center}

For categorical fields, the check requires exact agreement with the intended label. For value and answer fields, the check uses normalized or semantic equivalence rather than raw string identity, so answers such as ``7.5'' and ``7.5 hours per night'' are treated as the same value when the question asks for average sleep duration. The only semantic disagreements are two current answer cases where one annotator judged the current cue under specified for the original topic while the other accepted a broader paraphrase. The raw exact string scorer gives perfect agreement on state slot and update type, but lower exact agreement on value fields because one annotator often writes compact values and the other writes values with units or predicates.

\subsection{Answer Judge Reliability Check}
\label{app:judge-reliability}

Because the main tables use \texttt{llama3.3:70b} as an offline answer judge, we audit a random 100 output sample with human verification. The audit contains 50 LTP outputs and 50 LoCoMo outputs, sampled to cover base and +\ATMA rows and the judge's three labels: correct, partially correct, and incorrect. The human sheet hides the method name, source file, and 70B label. The annotator sees only the dataset, question, gold answer, alternate answer when present, model response, and retrieved evidence. Labels use the same three way rubric as the table judge, with semantic equivalence allowed for wording and date formats.

\begin{center}
\small
\begin{tabular}{@{}lrrrr@{}}
\toprule
\textbf{Scope} & \textbf{N} & \textbf{3-way agree.} & \textbf{3-way $\kappa$} & \textbf{Binary $\kappa$} \\
\midrule
All outputs & 100 & 80.0\% & 0.699 & 0.847 \\
LTP & 50 & 88.0\% & 0.820 & -- \\
LoCoMo & 50 & 72.0\% & 0.578 & -- \\
\bottomrule
\end{tabular}
\end{center}

The binary view maps correct and partially correct to acceptable and keeps incorrect separate. Under this view, the overall human--LLM agreement is 93.0\% with Cohen's $\kappa=0.847$. The remaining disagreements are concentrated around partial credit boundaries, especially relative time expressions in LoCoMo and answers that recover the main entity but omit a required qualifier. We therefore use the 70B judge as a reliable coarse answer correctness audit while treating fine grained partial credit as the noisier part of the evaluation.

\subsection{Metric Definitions}
\label{app:metric-definitions}

\noindent\textbf{LTP metrics.}
\bankcoexist measures whether the final bank contains matched records for both the current gold value and the old gold value. \bankrole measures whether those matched records also carry an inspectable state-role signal under the method's native interface. The role scorer uses explicit status fields and supersession, transition, or coexistence links when they exist. Timestamp-only ordering is not credited: a newer timestamp can help retrieval or QA, but it is not counted as a state role by itself. Scope updates can receive role credit when both values remain active under explicit coexistence or link evidence. Thus \bankcoexist measures value preservation, while \bankrole measures whether the preserved values are state typed. \evidsupport measures whether the retrieved evidence packet contains support for the answer requested by the query state. QA Acc. measures whether the final answer resolves the requested target state. Conflict and fact accuracy split the QA probes into state conflict probes and stable fact probes. Judge is the average 70B judge score for final answer quality under the LTP rubric.

\noindent\textbf{LoCoMo metrics.}
F1 and BLEU-1 are lexical overlap metrics against the benchmark gold answer. QA Acc. is a judged answer correctness score under a LoCoMo specific rubric. We keep the LoCoMo judge separate from the LTP judge because the benchmarks ask different questions. LTP evaluates controlled state resolution against target and alternate states. LoCoMo evaluates whether a long conversation answer semantically matches the benchmark gold answer. An evidence faithful abstention can be acceptable only for genuinely unanswerable adversarial questions. For answerable LoCoMo questions, abstaining because the retrieved evidence is insufficient is judged wrong.

\noindent\textbf{Rank.}
All reported metrics are treated as higher better except Rank. The LTP Rank column is the average rank across the seven completed LTP metrics in Table~\ref{tab:main-ltp-full}: \bankcoexist, \bankrole, \evidsupport, QA Acc., conflict accuracy, fact accuracy, and Judge. The LoCoMo Rank column is the average rank across the three average metrics in Table~\ref{tab:locomo-full-continued}: average F1, average BLEU-1, and average QA Acc. Ties receive the average of the tied rank positions induced by the displayed values. Rank is a compact summary for table reading; it is not used as a separate training or selection objective.

\subsection{Host-System Evaluation Policy}
\label{app:host-evaluation-policy}

\ATMA is evaluated as an overlay on a host memory system, not as a replacement for that system. A base row reports the host under the benchmark evaluation path. A +\ATMA row keeps the same host family and adds state aware bank labels, state aligned evidence construction, and answer side state serialization. This setup tests whether explicit state roles improve the host pipeline when useful evidence is available. It does not imply that \ATMA can recover evidence that the host never stores or retrieves.

Within each full comparison, the answer model, retrieval budget, and 70B judge are held fixed as far as the method interface allows. This isolates state coordination across bank, retrieval, and QA from changes to the generator or judge. When a method cannot complete a full run under the same evaluation scope, its corresponding cells are left blank.

\begin{table*}[!t]
\centering
\caption{Component ablation on LTP Profiles 000--002 under the same A-Mem host family and 70B judge. The first four columns mark which \ATMA components are enabled. Metrics follow the main LTP table format, with higher better; Pref. denotes strict preferred accuracy under the judge. The full \ATMA row is highlighted in light blue.}
\label{tab:level-ablation-profiles000-002}
\renewcommand\tabcolsep{2.25pt}
\renewcommand\arraystretch{1.12}
\footnotesize
\resizebox{\linewidth}{!}{%
\begin{tabular}{l|cccc|c|r|r|rrrrr}
\Xhline{1.2pt}
\rowcolor{atmaheader}
& \multicolumn{4}{c|}{\textbf{\ATMA Component}} & & \multicolumn{1}{c|}{\textbf{Bank Level}} & \multicolumn{1}{c|}{\textbf{Retrieval Level}} & \multicolumn{5}{c}{\textbf{QA Level}} \\
\rowcolor{atmaheader}
\multirow{-2}{*}{\textbf{Variant}} & \textbf{Sentry} & \textbf{Retrieval} & \textbf{Judge} & \textbf{QA Label} & \multirow{-2}{*}{\textbf{N}} & \textbf{\bankcoexist} & \textbf{\evidsupport} & \textbf{QA Acc.} & \textbf{Pref.} & \textbf{Conflict} & \textbf{Fact} & \textbf{Judge} \\
\Xhline{1.2pt}
A-Mem host & \nomark & \nomark & \nomark & \nomark & 240 & \textbf{1.000} & 0.917 & 0.825 & 0.567 & 0.825 & 0.808 & 0.706 \\
$-\textsc{Sentry}$ & \nomark & \yesmark & \yesmark & \yesmark & 240 & \textbf{1.000} & 0.917 & \underline{0.829} & 0.571 & \underline{0.850} & \underline{0.817} & 0.710 \\
$-\textsc{Retrieval Controller}$ & \yesmark & \nomark & \yesmark & \yesmark & 240 & \textbf{1.000} & 0.917 & 0.817 & \underline{0.583} & 0.833 & 0.800 & \underline{0.713} \\
$-\textsc{Judge}$ & \yesmark & \yesmark & \nomark & \yesmark & 240 & \textbf{1.000} & \underline{0.921} & 0.817 & 0.571 & 0.833 & 0.800 & 0.704 \\
$-\textsc{QA Label}$ & \yesmark & \yesmark & \yesmark & \nomark & 240 & \textbf{1.000} & \underline{0.921} & 0.825 & 0.579 & \underline{0.850} & 0.792 & 0.708 \\
\rowcolor{atmahlight} Full A-Mem+\ATMA & \yesmark & \yesmark & \yesmark & \yesmark & 240 & \textbf{1.000} & \textbf{0.958} & \textbf{0.883} & \textbf{0.638} & \textbf{0.858} & \textbf{0.825} & \textbf{0.759} \\
\Xhline{1.2pt}
\end{tabular}}
\end{table*}

\subsection{Implementation Details}
\label{app:implementation-details}

\noindent\textbf{Sentry.}
The deployed Sentry implementation uses \texttt{nomic-ai/nomic-embed-text-v1.5} as the SentenceTransformer backbone with \texttt{trust\_remote\_code=True}. The backbone output is passed to two MLP projection heads: a topic head with a ReLU hidden layer and a logic head with a Tanh hidden layer. The active gate thresholds are $\tau_{\mathrm{top}}=0.60$, $\tau_{\mathrm{log}}=0.45$, and $\tau_{\mathrm{gap}}=0.15$, so a pair is routed to the audit Judge only when topic similarity is high and either logic similarity is low or the topic logic gap is large. The source pool for this tier is the reconstructed profile interaction data: 50 profiles, 500 profile facts, and about 1,000 interaction fact examples. For Sentry training, each interaction contributes the same slot pair and one sampled irrelevant negative pair. The training script splits by interaction id, optimizes the supervised topic and logic MSE objective with AdamW, and selects \texttt{sentry\_best.pt} by validation loss. The active H200 setup uses the resulting Sentry checkpoint in the \texttt{atma-assets} directory.

\noindent\textbf{Audit Judge.}
There are two Judge surfaces in the codebase. The ATMA bank update path uses a compact audit model, while the tables use a separate 70B answer judge for evaluation. The audit Judge is based on \texttt{Qwen/Qwen2.5-3B-Instruct}. It is first warmed up with QLoRA SFT using CoT samples derived from \texttt{interactions\_v2.json} and \texttt{profiles\_v2.json}; the logged SFT run uses 700 English training items, reports train loss 0.6712 and eval loss 0.4039, and reaches 82.1\% accuracy, precision 0.8551, recall 0.7763, and F1 0.8138 on the aligned 151 item held out set. The SFT script uses 4 bit NF4 quantization, LoRA rank 64, LoRA alpha 16, dropout 0.05, batch size 4, gradient accumulation 4, learning rate $2\times10^{-4}$, and maximum sequence length 1024.

The GRPO stage starts from the merged SFT checkpoint and uses the CoT training data in the v6 format. Its default configuration uses LoRA rank 16, alpha 32, learning rate $2\times10^{-5}$, 200 steps, per device batch size 1, gradient accumulation 4, two generations per prompt, maximum completion length 400, and KL coefficient 0.04. The reward is rule based rather than learned. The format reward is discrete: 1.0 for a valid \texttt{<think>} block with the four required reasoning step labels, 0.5 for valid tags without all step labels, and 0 otherwise. Other reward terms add or subtract fixed values for verdict correctness, implicit violation success, length, tag cohesion, conclusion conciseness, prompt echoing, old memory preservation, and transition summary quality. The logged GRPO run reaches 87.4\% accuracy and F1 0.8774 on the same 151 item held out set, with implicit violation accuracy improving from 25.0\% for SFT to 75.0\%. The active H200 ablation manifest records \texttt{qwen3b-grpo-v6} as both the audit model and retrieve controller model. The reported table judge is instead \texttt{llama3.3:70b}; it is not the model used to write ATMA bank metadata.

\noindent\textbf{Query profiler.}
The query profiler is rule based. It counts lexical hints for historical state, current state, and change state. Change wins when change hints dominate, or when both old and current hints appear. Otherwise the profiler returns historical, current, or neutral by the larger hint count. It also records three boolean guards for negation, scope, and update wording. These features are passed to retrieval control and trace logging; no learned classifier or LLM call is used for query profiling.

\noindent\textbf{Retrieve controller.}
The retrieve controller uses \texttt{qwen3b-grpo-v6} by default through the Ollama/OpenAI compatible endpoint. The controller sees only a bounded shortlist from the candidate pool: at most 10 rows, each with content truncated to 220 characters. It may select at most 3 rows, clipped by the final top $k$, and the remaining packet positions are filled by the stable pre rank. The default generation cap is 320 tokens, timeout is 90 seconds, temperature is 0, and the active mode can be \texttt{shadow} or \texttt{online}. In shadow mode the controller output is logged but not applied. In online mode it is applied only when JSON parsing succeeds and the selected candidate ids are valid. The controller system prompt is:

\begin{tcolorbox}[notitle, sharp corners, colframe=gray, colback=white,
       boxrule=1.2pt, boxsep=0.5pt,
       title={Retrieve Controller Prompt},]\label{box:retrieve_controller_prompt}
       \footnotesize\ttfamily\raggedright
You are A-TMA's retrieve-level controller.\\
\smallskip
You do retrieval control only.\\
Do not answer the user question.\\
Do not use outside knowledge.\\
Use only the provided candidate memories.\\
\smallskip
Your job:\\
1. infer the requested state view: current, historical, change, or neutral;\\
2. select and rank only the highest-priority candidate evidence rows for state-correct evidence construction;\\
3. prefer direct state evidence over broad weak hops;\\
4. keep transition evidence only when it helps answer a change-related or ambiguity-sensitive query;\\
5. be concise: short reasons only;\\
6. return strict JSON only.
\end{tcolorbox}
The user prompt is a JSON object with \texttt{task=retrieve\_control}, the query, \texttt{top\_k}, \texttt{max\_selected}, the rule based query profile, the compact candidate list, the required output schema, and short instructions. The required output fields are \texttt{query\_mode}, \texttt{confidence}, \texttt{fallback\_recommended}, a \texttt{selected} list containing \texttt{candidate\_id}, rank, role, state fit, scope fit, change fit, noise risk, and a short reason, plus \texttt{packing\_hints} for transition evidence, contrast pairs, and weak hops.

\noindent\textbf{QA model, retrieval budget, and overhead.}
The full LTP script fixes the QA model to \texttt{qwen2.5:3b}, the memory model to \texttt{qwen2.5:3b} unless a host requires its own local model, the retrieval budget to top $k=5$, and the evaluation judge to \texttt{llama3.3:70b}. The H200 ablation manifest uses \texttt{qwen3:30b} for memory construction but keeps \texttt{qwen2.5:3b} for QA and \texttt{llama3.3:70b} for judging. The QA prompt prepends each retrieved row with one of five labels: current memory, historical memory, transition memory, transition linked memory, or retrieved memory, and then asks the model to answer only from the retrieved memories. This adds a short label and optional timestamp to each of at most five evidence rows. The retrieve controller adds one bounded JSON call over at most 10 truncated candidates and caps its output at 320 tokens. The 70B judge is an offline evaluation service and is excluded from deployment overhead. On the standalone 151 item audit diagnostic, Sentry runs at about 50 ms per pair, the GRPO Judge at about 8 seconds per escalated pair, and the cascade sends 23.2\% of pairs to the Judge, giving about 2 seconds average diagnostic latency. We do not use these diagnostic latencies as a main LTP or LoCoMo result.

\begin{atmaalgo}
\footnotesize
\newcommand{\algcomment}[1]{{\scriptsize\textcolor{lightblue}{// #1}}}
\refstepcounter{algorithm}\label{alg:atma_inference}
\noindent\rule{\columnwidth}{0.4pt}
\vspace{1pt}
\noindent\textbf{Algorithm \thealgorithm: \ATMA end to end inference}\\
\textbf{Input}: stream $M$, query $q$, host APIs $(B_H,R_H,G_H)$\\
\textbf{Output}: answer $\hat{a}$, trace $T_q$\\
\noindent\rule{\columnwidth}{0.4pt}
\begin{tabular}{@{}r p{0.86\columnwidth}@{}}
\multicolumn{2}{@{}l}{\textit{Bank side:}}\\
1 & $B\leftarrow\operatorname{InitStateBank}(B_H)$ \quad \algcomment{preserve host records}\\
2 & \textbf{for} $m_t\in M$ \textbf{do}\\
3 & \quad $\mathcal{P}_t\leftarrow R_H(m_t,B)\cup\operatorname{Link}(m_t,B)$ \quad \algcomment{collect possible prior states}\\
4 & \quad $\mathcal{A}_t\leftarrow\{r_i\in\mathcal{P}_t:S(r_i,m_t)=1\}$ \quad \algcomment{keep only risky pairs}\\
5 & \quad $d_t\leftarrow J(\mathcal{A}_t,m_t)$ \quad \algcomment{skip Judge when no conflict exists}\\
6 & \quad $B\leftarrow\operatorname{Commit}(B,m_t,d_t)$ \quad \algcomment{write status labels and state links}\\
7 & \textbf{end for}\\[1pt]
\multicolumn{2}{@{}l}{\textit{Retrieval side:}}\\
8 & $\psi(q),v_q\leftarrow\operatorname{Profile}(q)$ \quad \algcomment{detect requested state view}\\
9 & $A_q\leftarrow R_H(q,B)$ \quad \algcomment{reuse the host seed retriever}\\
10 & $\mathcal{C}_q\leftarrow\operatorname{Expand}(A_q,\mathrm{sup},\mathrm{tran},\mathrm{nbr})$ \quad \algcomment{add linked state evidence}\\
11 & $E_q^0\leftarrow\operatorname{TopK}(\operatorname{Dedup}(\mathcal{C}_q),k)$\\
12 & $E_q\leftarrow\operatorname{Select}(F_\eta(q,E_q^0),E_q^0)$ \quad \algcomment{fall back to stable pre-rank}\\[1pt]
\multicolumn{2}{@{}l}{\textit{QA side:}}\\
13 & $\widetilde{E}_q\leftarrow\operatorname{Serialize}(E_q,\lambda)$ \quad \algcomment{expose current, historical, transition roles}\\
14 & $\hat{a}\leftarrow G_H(\operatorname{Prompt}(q,v_q,\widetilde{E}_q))$ \quad \algcomment{force the answer to follow the requested view}\\
15 & $T_q\leftarrow(\psi(q),\mathcal{C}_q,E_q,\widetilde{E}_q,\hat{a})$\\
16 & \textbf{return} $\hat{a},T_q$\\
\end{tabular}
\noindent\rule{\columnwidth}{0.4pt}
\end{atmaalgo}

\subsection{Inference Procedure}
\label{app:inference-procedure}

Algorithm~\ref{alg:atma_inference} summarizes the inference path. The host keeps its native memory insertion, retrieval, and generation interfaces. \ATMA adds state maintenance, packet construction, and trace aware answer conditioning.

The trace $T_q$ matches the decoupled evaluation used in the paper. Bank errors appear as missing or wrongly labeled state records. Retrieval errors appear as missing or wrongly ordered packet evidence. QA errors appear when labeled evidence is present but the answer resolves the wrong state.

\subsection{Level Ablation on LTP Profiles 000--002}
\label{app:level-ablation-profiles000-002}

Table~\ref{tab:level-ablation-profiles000-002} reports a matched component ablation on LTP Profiles 000--002. The run covers 120 state units and 240 probes. All variants use the same A-Mem host family, QA model, and 70B judge. The full overlay is competitive, while the best preferred accuracy and average judge score in this subset come from the variant without Sentry.

For this component ablation, the bank column is the stricter ATMA-internal status diagnostic rather than the cross-method personalized scorer used in the full main table. All \ATMA bank variants reach the same score of 0.500 on this subset, while the A-Mem host row has no explicit status signal under that diagnostic. Because the \ATMA variants share the same bank-side value, the QA differences mainly isolate retrieval control and answer side labels once the controlled bank-side signal is held fixed.

The ablation shows small and mixed component effects. Removing the retrieval controller lowers QA Acc. from 0.829 to 0.817, while evidence support stays at 0.917 and the average judge score remains close. Removing QA state labels changes the tradeoff rather than causing a uniform collapse: conflict accuracy rises to 0.858, but fact recall falls to 0.792. Removing Sentry gives the highest preferred accuracy at 0.588 and judge score at 0.719, suggesting that the lightweight bank proposal gate is not the bottleneck for these three profiles.

\subsection{Qualitative Case Study}
\label{app:qualitative-case-study}

The following case studies list the question, target state, retrieved evidence, QA output, and judge result for two pipeline positions where the component effect is directly visible: answer-side labels and retrieval control.

\begin{tcolorbox}[notitle, sharp corners, colframe=gray, colback=white,
       boxrule=1.2pt, boxsep=0.5pt,
       title={Case 1: QA state labels},]
       \footnotesize\raggedright
\textbf{Question.} ``Before the update, how many hours per night did you sleep on average?''\\
\textbf{Target / alternate.} ``5'' / ``7.''\\
\textbf{Retrieved evidence (top rows).}
\begin{enumerate}[leftmargin=1.5em,itemsep=0pt,topsep=2pt]
\item ``Speaker Caroline says : I sleep about 7 hours per night on average.''
\item ``Speaker Caroline says : I sleep about 5 hours per night on average''
\item ``Speaker Caroline says : Wow! Did you see that band?''
\item ``Speaker Melanie says : Wow, looks awesome! Did you join in?''
\end{enumerate}
\textbf{QA serialization.} Full \ATMA passes the same evidence with the requested historical state view, while the no-label variant passes the rows without explicit state roles.\\
\textbf{QA responses.} Full \ATMA: ``Before the update, Caroline slept about 5 hours per night on average.'' Judge: correct, 1.0. No QA label: ``7 hours.'' Judge: wrong, 0.0.
\end{tcolorbox}

\begin{tcolorbox}[notitle, sharp corners, colframe=gray, colback=white,
       boxrule=1.2pt, boxsep=0.5pt,
       title={Case 2: Retrieve controller},]
       \footnotesize\raggedright
\textbf{Question.} ``Before the update, what was the value for i drive about?''\\
\textbf{Target / alternate.} ``3000'' / ``4500.''\\
\textbf{Retrieved evidence without controller (top rows).}
\begin{enumerate}[leftmargin=1.5em,itemsep=0pt,topsep=2pt]
\item ``Speaker Caroline says : Sounds fun! What was the best part? Do you do it often with the kids?''
\item ``Speaker Melanie says : Wow, sounds amazing! What was the event like? Those posters are great!''
\item ``Speaker Caroline says : I drive about 3000 miles per year in my 14-year-old car''
\item ``Speaker Caroline says : I drive about 4500 miles per year in my 14-year-old car.''
\end{enumerate}
\textbf{Retrieved evidence with controller (top rows).}
\begin{enumerate}[leftmargin=1.5em,itemsep=0pt,topsep=2pt]
\item ``Speaker Caroline says : I drive about 3000 miles per year in my 14-year-old car''
\item ``Speaker Caroline says : Sounds fun! What was the best part? Do you do it often with the kids?''
\item ``Speaker Melanie says : Wow, sounds amazing! What was the event like? Those posters are great!''
\item ``Speaker Caroline says : I drive about 4500 miles per year in my 14-year-old car.''
\end{enumerate}
\textbf{QA responses.} Shadow baseline: ``4500 miles per year.'' Judge: wrong, 0.0. Online controller: ``3000 miles per year.'' Judge: correct, 1.0.\\
\textbf{Run note.} QA model: \texttt{qwen2.5:3b}; judge: \texttt{llama3.3:70b}.
\end{tcolorbox}

The first trace isolates answer-side serialization: the retrieved rows are unchanged, but explicit state roles prevent the answer model from treating the current value as the historical answer. The second trace isolates retrieval ordering: the controller moves the requested historical value ahead of the conflicting current value, which changes the final answer from the wrong state to the requested state.

\end{document}